\documentclass[10pt,twocolumn,letterpaper]{article}

\usepackage{cvpr}
\usepackage{times}
\usepackage{epsfig}
\usepackage{graphicx}
\usepackage{amsmath}
\usepackage{amssymb}
\usepackage{paralist}
\usepackage[top=2cm, bottom=2cm, left=2cm, right=2cm]{geometry}  

\usepackage[pagebackref=true,breaklinks=true,letterpaper=true,colorlinks,bookmarks=false,citecolor=blue,linkcolor=blue]{hyperref}

\cvprfinalcopy 

\def\cvprPaperID{****} 

\ifcvprfinal\fi
\begin{document}
	
	\title{Learning to Segment Instances in Videos with Spatial Propagation Network}
	
	\author{Jingchun Cheng$^{1,2}$
		\hspace{0.15in} Sifei Liu$^{2}$ \hspace{0.15in} Yi-Hsuan Tsai$^{2}$ \hspace{0.15in} Wei-Chih Hung$^{2}$
		\hspace{0.15in} Shalini De Mello$^{3}$ \\
		\hspace{0.15in} Jinwei Gu$^{3}$ \hspace{0.15in} Jan Kautz$^{3}$ \hspace{0.15in} Shengjin Wang$^{1}$ 
		\hspace{0.15in} Ming-Hsuan Yang$^{2}$
		\vspace{1mm} \\
		\hspace{0.15in} $^{1}$Tsinghua University \hspace{0.15in} $^{2}$University of California, Merced
		\hspace{0.15in} $^{3}$NVIDIA Research \\
	}

	\maketitle
	
	\begin{abstract}
		We propose a deep learning-based framework for instance-level object segmentation. Our method mainly consists of three steps.
		First, We train a generic model based on ResNet-101 for foreground/background segmentations.
		Second, based on this generic model, we fine-tune it to learn instance-level models and segment individual objects by using augmented object annotations in first frames of test videos.
		To distinguish different instances in the same video,
		we compute a pixel-level score map for each object from these instance-level models.
		Each score map indicates the objectness likelihood and is only computed within the foreground mask obtained in the first step.
		To further refine this per frame score map, we learn a spatial propagation network.
		This network aims to learn how to propagate a coarse segmentation mask spatially based on the pairwise similarities in each frame.
		In addition, we apply a filter on the refined score map that aims to recognize the best connected region using spatial and temporal consistencies in the video.
		Finally, we decide the instance-level object segmentation in each video by comparing score maps of different instances.
	\end{abstract}
	
	\vspace{-3mm}
	\section{Introduction}
	In this work, we focus on the problem of multiple instance segmentation in videos. Specifically, given each object mask in the first frame, we seek to predict segmentations for this instance throughout the video sequence.
	The task is challenging when dealing with non-rigid objects (e.g., human, animals) because these objects often have their individual movements with various perspectives, poses.
	Occlusions can also pose significant challenges for tracking based methods since the foreground objects could be fully occluded in some frames. With the multiple instance setting, occlusions between different instances also introduce further difficulties to keep tracking each instance separately.
	
	Most state-of-the-art approaches tackle the problem with convolutional neural networks (CNNs)~\cite{khoreva2016learning,caelles2016one}.
	Intuitively, CNNs are trained to output the foreground/background segmentation maps following the structure of the fully convolution networks (FCN)~\cite{long2015fully} for every frame in the video sequence.
	In the unsupervised setting, a general foreground model is learned with the training set. Under the semi-supervised setting, one can further fine-tune the model using the segmentation mask in the first frame of the test video to focus on the particular foreground region.
	To extend this pipeline to the multiple instance setting, we decompose this task into foreground segmentation and instance recognition. While foreground segmentation can be trained on the training set with respect to all foreground objects, instance recognition can be trained on each specific instance to separate the foreground mask into multiple instances.
	
	We observe that similar to most FCN based segmentation methods~\cite{long2015fully, chen2014semantic,zheng2015conditional}, segments generated by the network are often not aligned to the actual object boundaries because of the pooling operations during forward propagation.
	To address this issue, many existing methods on image-level semantic segmentation task apply the conditional random field (CRF) as the post-processing module to refine object boundaries~\cite{chen2014semantic, zheng2015conditional}.
	However, densely connected CRF requires sophisticated designs of potential functions and fine-tuned hyper-parameters. There are end-to-end trainable CRFs such as \cite{zheng2015conditional}, but they often introduce much memory and computational overhead.

\begin{figure*}
		\begin{center}
			\includegraphics[width=1\linewidth]{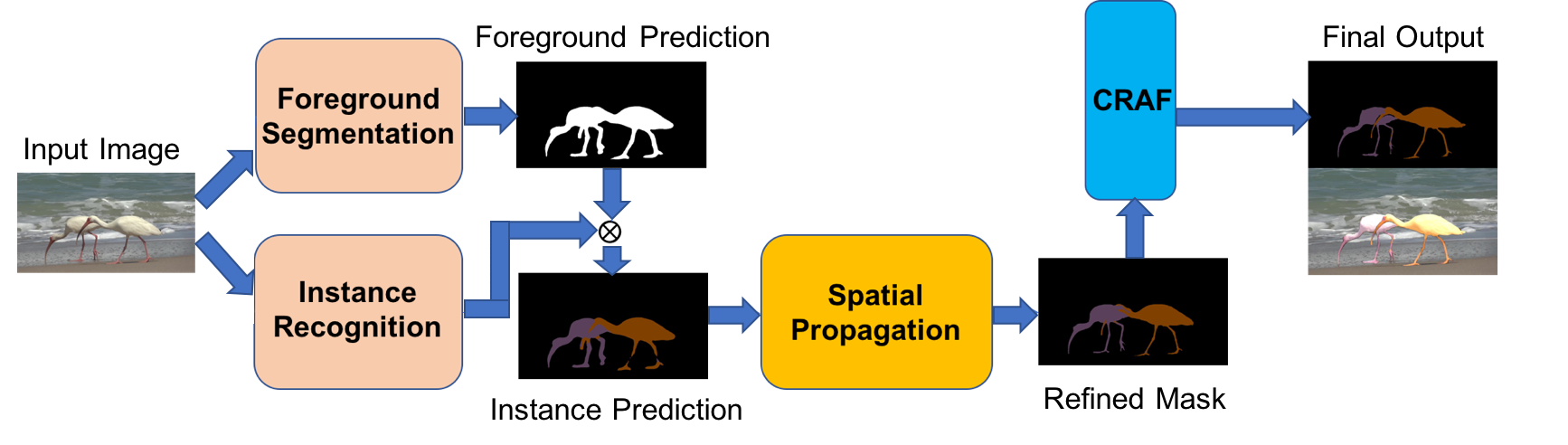}
		\end{center}
		\vspace{-3mm}
		\caption{Framework of the proposed method.}
		\vspace{-3mm}
		\label{fig:framework}
	\end{figure*}

	To address this issue, we model the boundary refinement task as a spatial propagation problem with pixel-wise affinity prediction. Specifically, we propose a spatial propagation network (SPN) that propagates the segmentation probabilities using the learned pixel-wise affinity as guidance with a linear 2D propagation module.
	To further refine the segments that are not consistent in the temporal domain, we propose the connected region-aware filter (CRAF) to eliminate inconsistent labels.
	Figure \ref{fig:framework} illustrates the overview of the proposed algorithm.
	
	To evaluate the proposed methods, We carry out extensive experiments and ablation studies on the DAVIS 2017 challenge dataset~\cite{Pont-Tuset_arXiv_2017}. We show that the proposed SPN improves the object boundaries, while the proposed CRAF eliminates segments with inconsistent instance labels.
	
	The contributions of this work are summarized below:
	\begin{compactitem}
		\item We extend the segmentation network to handle multiple instances simultaneously by decomposing the task into foreground segmentation and instance recognition.
		\item We propose the spatial propagation network to refine object segments through learning the spatial affinity.
		\item We develop the connected region-aware filter to eliminate inconsistent segments. 
	\end{compactitem}
	
	\section{Related Work}
	
	{\flushleft {\bf Video Object Segmentation.}}
	There are two problem settings for video object segmentation: unsupervised and semi-supervised ones.
	Unsupervised methods aim to segment foreground objects without any knowledge of the object during testing (e.g., an initial object mask).
	Several methods have been proposed to generate object segmentation via superpixel \cite{grundmann2010efficient, xu2012streaming, galasso2012video}, saliency \cite{rahtu2010segmenting, faktor2014video, Wang_CVPR_2015}, or optical flow \cite{Brox_ECCV_2010, Papazoglou_ICCV_2013}.
	To incorporate higher level information such as objectness, object proposals are used to track object segments and generate consistent regions through the video~\cite{Lee_ICCV_2011, Li_ICCV_2013}.
	However, these methods usually have heavy computational loads to generate region proposals and associate thousands of segments, making such approaches only feasible to offline applications.

	Semi-supervised approaches \cite{Galasso_ICCV_2013, Zhang_CVPR_2013, Nagaraja_ICCV_2015} assume an object mask in the first frame is known, and the objective is to track the object mask throughout the video.
	To achieve this, existing approaches focus on propagating superpixels \cite{Jain_ECCV_2014}, constructing graphical models \cite{marki2016bilateral, tsai2016video} or utilizing object proposals \cite{perazzi2015fully}.
	Recently, CNN based methods \cite{khoreva2016learning, caelles2016one} combine offline and online training on static images.

	{\flushleft {\bf Instance Segmentation.}}
	Our work is also related to instance segmentation in the image level, especially to the subtasks including occlusion handling and boundary refinement.
	Most state-of-the-art approaches tackle this task using region proposals~\cite{ren2015faster} followed by object mask prediction.
	In \cite{dai2015instance}, a multiple stage network is used to predict bounding box proposals, mask proposals, and class score iteratively.
	However, the performance of instance segmentation often suffers from heavy occlusions.
	To handle occlusions, one can apply dense CRF on the patch level to generate instance masks~\cite{zhang2015instance}. In a non-parametric approach, exemplar segments from the training set are utilized to help the occlusion handling~\cite{chen2015multi}.

	To further obtain boundary accuracies between instances,
	probabilistic models can be applied as a
	post-processing module to refine object boundaries and enforce spatial smoothness.
	A straightforward approach is to apply fully-connected CRF in the testing phase~\cite{chen2014semantic}.
	In ~\cite{zheng2015conditional}, CRF is formulated as an RNN, resulting in an end-to-end trainable network with the spatial smoothness constraint.
	In our work, the proposed SPN directly learns the pixel affinity in an end-to-end manner from the data itself. It results in a light-weighted, computationally efficient refinement module.
	
	\begin{figure}
		\begin{center}
			\includegraphics[width=1\linewidth]{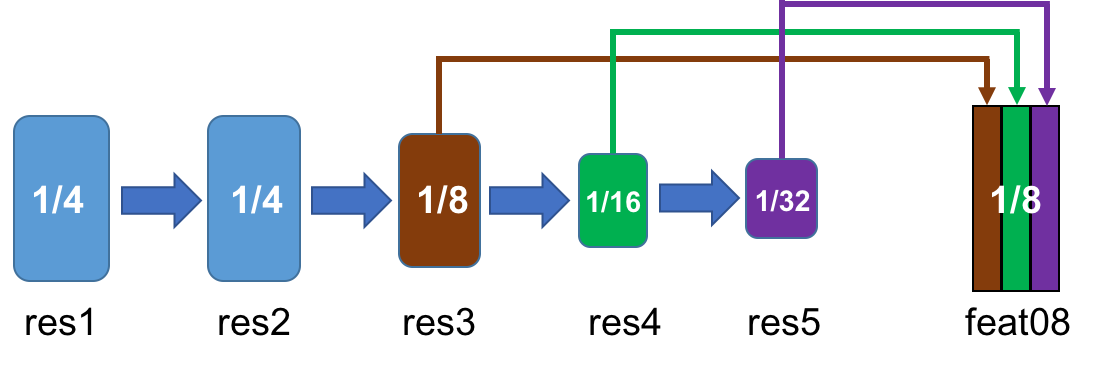}
		\end{center}
		\vspace{-3mm}
		\caption{Network architecture for foreground segmentation and instance recognition. The first five blocks (res1 to res5) are adapted from the ResNet-101.}
		\vspace{-5mm}
		\label{fig:resnet}
	\end{figure}
	
	\section{Learning to Segment Instances}
	Given the object mask of the first frame at the instance level, our goal is to segment this instance throughout the entire video.
	Toward this end, we first train a generic foreground/background segmentation model to localize objects and then fine-tune this generic model to learn instance-level models.

	\subsection{Foreground Segmentation}
	Inspired by the effectiveness of FCN in image segmentation \cite{long2015fully} and the deep structure in image classification \cite{he2016deep},
	we construct our foreground/background segmentation network based on the ResNet-101 architecture \cite{he2016deep} with modifications for pixel-wise segmentation predictions as follows:
	1) the fully-connected layer for classification is removed, and
	2) features of convolution modules in different levels are fused together for obtaining more details during up-sampling.

	The ResNet-101 has five convolution modules, where each of them consists of several convolutional layers.
	Specifically, we draw feature maps from the 3-th to 5-th convolution modules after the pooling operations, where these maps are with sizes of 1/8, 1/16, 1/32 of the input image size, respectively.
	Then these maps are up-sampled and concatenated together for predicting the final output (see Figure \ref{fig:resnet} for an illustration).

	A pixel-wise cross-entropy loss with the softmax function $\mathbb{E}$ is used during optimization.
	To overcome imbalanced pixel numbers between foreground and background regions, we use the weighted version as adopted in \cite{Xie_ICCV_2015}, and the loss function is defined as:
	
	\begin{footnotesize}
		\begin{align}
		\mathcal{L}_s(I_t) = - (1-w)\sum_{i,j\in fg} \log \mathbb{E}(y_{ij} = 1;\theta) \notag \\
		- w\sum_{i,j\in bg} \log \mathbb{E}(y_{ij} = 0;\theta),
		\label{eq:seg}
		\end{align}

	\end{footnotesize}
	
	\noindent
	where $i, j$ denotes the pixel location of foreground $fg$ and background $bg$, $y_{ij}$ denotes the binary prediction of each pixel of the input image $I$ at frame $t$, and $w$ is computed as the foreground-background pixel-number ratio.
	\subsection{Instance-level Recognition}
	After discovering foreground segmentations, the next step is to further segment instance-level objects.
	To achieve this, we adopt the same model and loss function in \eqref{eq:seg} for foreground segmentation, and fine-tune it for instance segmentation.
	As a result, for each instance we train a model, where the softmax function in \eqref{eq:seg} has two channels for the object instance and the background.

	Since each video may have multiple object instances and different instance-level models may not agree to each other, we develop a method to solve such confusions, e.g., two objects are close to each other.
	We compute a pixel-wise score map for each object from the output of the instance-level model, in which this score map indicates the likelihood of instance segmentation.
	To take advantage of the foreground model and reduce noisy segments, we also enforce the score map being non-zero only within the foreground segmentation.
	Once we have score maps from different instances, we determine the final instance-level segmentation results by labeling the one with the maximum score for each pixel. 
	
	\subsection{Network Implementation and Training}
	To train the foreground generic model, we first use annotations from the DAVIS training set, and then fine-tune on foreground masks with augmentations in the first frame of the DAVIS test set.
	When training the foreground generic model, we use weights from ResNet-101 \cite{he2016deep} as initializations. We use stochastic gradient descent (SGD) optimizer with batch size 1 and learning rate 1e-8 for 100,000 iterations.
	During training instance-level models, we then fine-tune this generic model by using augmented instance-level annotations on the test set.
	For instance-level models, we use batch size 1, starting from learning rate 1e-8 and decreasing it by half for every 10,000 iterations with a total of 30,000 iterations.
	Since the number of total training samples is relatively small, we adopt affine transformations (i.e., shifting, rotation, flip) to generate one thousand samples for each frame.

	\section{Mask Refinement}
	In this section, we refine the mask in a frame-wise manner. This is done by a spatial propagation network (SPN) that improves the object shapes from a coarse shape to a finer one under the guidance of the original frame, and a connected region-aware filter (CRAF) that eliminates inconsistent regions.
	We note that these two refinement processes are independent to instances, in which a learned SPN can be applied to any instances.
	\subsection{Spatial Propagation Network}
	The SPN contains a deep CNN that learns the affinity entities, and a spatial linear propagation module that refines a coarse mask.
	The coarse mask is refined under the guidance of the affinity, the learned pairwise relations, for any pairs of pixels.   
	All modules are differentiable and jointly trained using the SGD method. 
	The spatial linear propagation module is computationally efficient for inference due to the linear time complexity of the recurrent architecture.
	
	{\flushleft {\bf Method.}}
	%
	The SPN contains a propagation module that builds a learnable graph through linear propagation over a 2D map.
	Let $\mathbf{H}\in\mathbb{R}^{m\times n}$ denotes a propagation hidden layer on top of a $m\times n$ feature map, $h_{ij}$ and $x_{ij}$ be the pixel at $\left(i,j\right)$ for the hidden layer and the feature map, respectively.
	We use $\left\lbrace p_{ij}^{K}\right\rbrace_{K\in\mathbb{N}\left(ij\right)}$ to represent a group of weights for $\left(i,j\right)$, where $K$ is a neighboring coordinate of $\left(i,j\right)$, denoted as $\mathbb{N}\left(ij\right)$.
	The 2D linear propagation along one direction (e.g., left-to-right) is:
	%
	\begin{equation}
	h_{ij} = \left( 1-\sum_{K\in\mathbb{N}\left(ij\right)}p_{ij}^{K}\right) x_{ij} + \sum_{K\in\mathbb{N}\left(ij\right)}p_{ij}^{K}h_{K},
	\label{eq:2rnn-spatial}
	\end{equation}
	where $h_{K}$ is an adjacent pixel of $\left(i,j\right)$ in the hidden layer.
	Taking the example of the left-to-right direction, the neighborhood $\mathbb{N}\left(ij\right)$ contains three nodes $\left\lbrace\left( i-1,j-1\right);\left( i-1,j\right);\left( i-1,j+1\right)  \right\rbrace $ from the previous column.
	Each $p_{ij}^{K}$ represents the weight between two adjacent pixels.
	In this way, one direction of propagation enables each pixel to receive the information from a triangular 2D plane, where the integration of four different directions (e.g., top-to-bottom and the other two with the reverse directions) enables it to receive information from all the other pixels of the image/feature map.
	The propagation in~\eqref{eq:2rnn-spatial} is performed as column-wise transitions, 
	which can be expressed by the following linear operation:
	%
	\begin{equation}
	H_{i} = \left( 1-P_{i-1,i}\right) X_{i} + P_{i-1,i}H_{i-1},
	\label{eq:2rnn-matrix}
	\end{equation}
	where $H_i$ and $X_i$ are the $i$-th column for the hidden layer and the feature map, $P_{i-1,i}\in\mathbb{R}^{n\times n}$ is a linear transition matrix, and the key factor for the transport of information from column $i-1$ to $i$.
	Corresponding to the three-neighbor connection, $P_{i-1,i}\in\mathbb{R}^{n\times n}$ is a tridiagonal matrix, a simple form whose system stability can be easily controlled through 
	the Gershgorin's theorem~\cite{highRNN}.
	%
	In the back propagation pass, the derivative $\sigma_{i+1,i}$ with respect to $P_{i+1,i}$ is
	%
	\begin{equation}
	\sigma_{i+1,i} = \theta_{i}\cdot\left(H_{i+1}-X_{i} \right),
	\label{eq:bp-matrix}
	\end{equation}
	where $\theta_{i}$ is the error for $H_{i}$ flowing back from the top layer.
	
	We use the guidance network, a regular deep CNN with symmetric downsample and upsample parts, to output all the elements of $P_{i-1,i}$. 
	The error signal flows in the reverse direction along the hidden layer and then propagates to the guidance network 
	such that the entire network can be trained end-to-end.
	
	{\flushleft {\bf Network Implementation.}}
	%
	We describe the implementation of the SPN, which contains two separate branches: 1) a deep CNN based guidance network that outputs all elements of the transformation matrix, and 2) the linear propagation module that outputs the refined segmentations.
	In this work, we use the VGG-16~\cite{Simonyan_CoRR_2015} pre-trained network from the \textit{conv1} to \textit{pool5} as the downsampling part of the guidance network.
	The upsampling part adopts the exactly symmetric architecture and is learned from scratch.
	The layers with the same dimension between the downsampling and upsampling part are connected with skipped links to leverage the features of different levels.

	The guidance network typically takes in RGB images.  
	It outputs all the connection weights w.r.t each pixel, where each has $3\times 4$ parameters to learn. 
	The propagation module takes a coarse segmentation mask produced by the previous step, and the weights generated by the guidance network.
	Suppose that we have a map of size $n\times n\times c$ that inputs into the propagation module, the guidance network needs to output a weight map with the dimensions of $n\times n\times c\times \left( 3\times 4\right) $,
	i.e., each pixel in the input map is paired with $3$ scalar weights per direction, and $4$ directions in total.
	The propagation module contains $4$ independent hidden layers for different directions, where each layer combines the input map with its respective weight map using \eqref{eq:2rnn-spatial}.
	%
	%
	Similar to~\cite{liu2016learningRecursive}, we use the node-wise max-pooling to integrate the hidden layers and obtain the final propagation result.
	
	{\flushleft {\bf Network Training.}}
	%
	There are two requirements to train the SPN.
	First, the SPN processes the two-class mask refinement.
	Second, for each training image with the ground truth annotation, a coarse mask is required.
	Therefore, we train our SPN on the training set of PASCAL VOC 2012~\cite{everingham2015pascal}, where the coarse mask is generated by the FCN~\cite{long2015fully}.
	For each image, we randomly pick a valid label according to the annotations, while treating all the other pixels as the background, in order to generate a two-class training sample out of the original 21 classes.
	During training, we randomly crop $256\times 256$ square patches from the image, the binary label, and the coarse mask.
	We note that there is only one single SPN as a general refinement module, no finetuning is carried out on any frame from Davis 2017 dataset.

	\begin{figure}
		\begin{center}
			\includegraphics[width=1\linewidth]{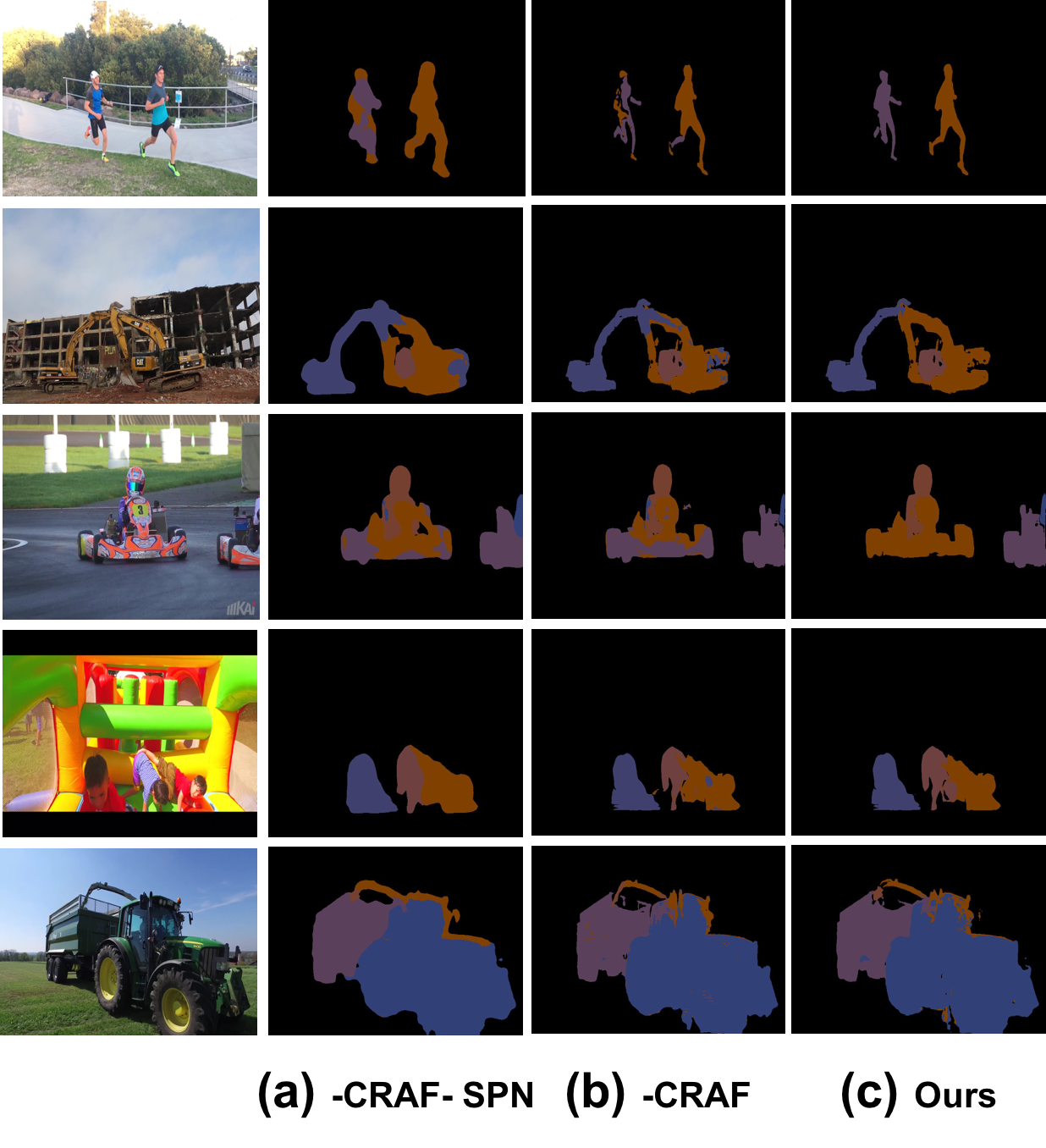}
		\end{center}
		\vspace{-3mm}
		\caption{Segmentation results after applying each step. (a) is the result without using any refinements, while (b) and (c) are the results without using CRAF and our final outputs, respectively.}
		\label{fig:figs_ablation}
	\end{figure}
    
	\subsection{Connected Region-aware Filter} 
	After applying the SPN, we observe that there exit many confusions between instances since we only do the segmentation in a one-shot manner without considering temporal information in other frames.
	To improve the instance mask, we use a connected region-aware filter (CRAF), which considers the consistency between two frames and helps rectify some instance confusions (see column (b) and (c) in Figure \ref{fig:figs_ablation} for illustrations).
	
	In CRAF, we select a best connected region for each object using the following criteria. 
	
	{\flushleft {\bf Step 1.}}
	For an object $i$, we extract connected regions ($CR_1, CR_2, ...$) on its score map after applying the SPN. 
	Then a jaccard similarity ($J_1, J_2, ...$) of each connected region compared to the region in the previous frame is computed, as well as the area of each region ($A_1, A_2, ...$).
	If $J_m = max(J_1, J_2, ...)$ and $A_m/max(A_1, A_2, ...)>\alpha$, we pick $CR_m$ as the best connected region for object $i$, denoted by $CRS_i$.
	If there is more than one jaccard similarities equal to the maximum value, we choose the connected region with the largest area to be $CRS_i$.
	
	{\flushleft {\bf Step 2.}}
	We calculate the coverage rate between $CRS$ of every two objects using the following formula: 
	\begin{equation}
	Coverage(i, j) = （|CRS_i \cap CRS_j|）/ |CRS_j|.
	\label{eq:Coverage}
	\end{equation}
	If $Coverage(i,j) > \beta$, we remove the regions of $CRS_j$ within $CRS_i$, i.e., $CRS_i = CRS_i - CRS_j$.
	
	{\flushleft {\bf Step 3.}}
	For an object $i$, if the region of $CRS_i < \gamma$, we check all the connected regions that are not selected as $CRS$. Then, if one of these regions has a jaccard similarity greater than $\delta$, we pick this connected region as $CRS_i$ and repeat Step 2.

	In this work, $\alpha, \beta, \gamma, \delta$ are empirically set as 0.2, 0.9, 0.1, 0.4 respectively.
	Note that, we set the scores outside each selected connected region as zero. To obtain the final instance segmentation, we determine the label for each pixel by considering the maximum score from all instances.

	\section{Experiments} 
	\subsection{Dataset and Evaluation Metrics}
	The DAVIS benchmark \cite{Pont-Tuset_arXiv_2017,Perazzi2016} is a recently-released high-quality video object segmentation dataset. 
	It consists of videos with multiple objects and instance-level pixel-wise annotations. In total, there are 150 sequences (60 in training set, 30 in each of the validation, test-dev and test-challenge sets), with 10459 annotated frames and 376 objects.
	
	We first use the training set to train our models and evaluate on the validation set.
	The best models are then trained on training and validation set and tested on the 2017 DAVIS Challenge for competition.
	The challenge uses the mean of region similarity (J mean) and contour accuracy (F mean) over all object instances as the performance metrics.
	The same algorithm (evaluation code from the DAVIS website) is used in the validation set to validate our method.
	
	\subsection{Comparisons of Instance-level Recognition}
	We compare two different models for instance-level segmentation: per-video and per-object settings.
	Initialized from weights of the foreground model, the per-video object recognition network has a softmax layer with a N+1 dimensional score map as the output, where N denotes the number of objects in the video with an additional one for the background.
	Each score map denotes the probability of one object. 
	In prediction, the pixel that has the maximum score is considered as the label of that pixel.

	For the per-object model, only one object is considered as the foreground each time.
	The network for each per-object model has a 2-dimensional output that contains the background and one object instance.
	In prediction, score maps of different objects are concatenated together, and
	if the maximum score is below 0.5, the pixel belongs to the background. 
	Otherwise, the object label of each pixel is determined by the instance with the maximum score.

	Comparisons of these two methods are shown in Table \ref{tab:per-vid vs per-obj}, where the models are trained on the training set and tested on the validation set.
	The results show that the per-object model outperforms the per-video model by 2.1\% in Global Mean.
	Therefore, we choose to use the per-object model for the following experiments.
	
	\begin{table}[t]\scriptsize
		\caption{Comparisons of instance segmentation models.}
		\begin{center}
			\small
			\centering
			\renewcommand{\arraystretch}{1.2}
			\setlength{\tabcolsep}{1.8pt}
			\begin{tabular}{|l|ccccc|}
				\hline
				Method                           & Global Mean & J Mean    & J Recall     & F mean  &   F Recall  \\
				\hline
				Per-video model      &0.460        &0.442      &0.513         &0.478    &0.501\\
				Per-object model     &0.481        &0.457      &0.536         &0.504    &0.526\\
				\hline
			\end{tabular}
		\end{center}
		
		\label{tab:per-vid vs per-obj}
		\vspace{-3mm}
	\end{table}
	
	\noindent

	\subsection{Ablation Study}
	To analyze the necessity and importance of each step in our proposed method, we carry out extensive ablation studies on the validation set. Results are summarized in Table \ref{tab:ablation}.
	We validate our method by comparing our final results to the ones without fine-tuning ({\bf -FT}), spatial propagation network ({\bf -SPN}), and connected region-aware filter ({\bf -CRAF}).
	The detailed settings are as follows:
	
	\noindent
	{\bf -FT}: train the foreground segmentation model without fine-tuning on first frames.
	
	\noindent
	{\bf -CRAF}: apply the SPN after instance segmentation without the use of the connected region-aware filter.
	
	\noindent
	{\bf -SPN}: the results from instance segmentation network without using spatial propagation network for refinement.

	Table \ref{tab:ablation} shows that the SPN and CRAF post-processing steps play an important role in generating better results, and improve the Global mean by 5.6\% and 3.9\% respectively.
	It also demonstrates that the foreground mask prediction network needs a fine-tuning step on the first frame for more accurate segmentations on the specific object (-CRAF-SPN vs. -CRAF-SPN-FT).
	Some example results after applying different steps are shown in Figure \ref{fig:figs_ablation}.

	\begin{table}[t]\scriptsize
		\caption{Ablation study on the DAVIS validation set.
			We show comparisons of our results with different components removed, i.e., foreground model fine-tuning (FT), spatial propagation network (SPN), connected region-aware filter (CDAF).}
		\begin{center}
			\small
			\centering
			\renewcommand{\arraystretch}{1.5}
			\setlength{\tabcolsep}{2.3pt}
			\begin{tabular}{|l|cccc|}
				\hline
				Method                          & Ours    & -CRAF      & -CRAF-SPN  &   -CRAF-SPN-FT   \\
				\hline
				J Mean              &0.540    &0.506       &0.457       &0.442\\
				J Recall            &0.601    &0.582       &0.536       &0.528\\
				\hline
				F Mean              &0.611    &0.568       &0.504       &0.453\\
				F Recall            &0.683    &0.602       &0.526       &0.501\\
				\hline
				Global Mean         &0.576    &0.537       &0.481       &0.448 \\
				\hline
			\end{tabular}
		\end{center}
		
		\label{tab:ablation}
	\end{table}
	
	\begin{table}[t]\scriptsize
		\caption{Runtime Analysis.
			We show the runtime in foreground segmentation (FS), instance-level recognition (IR), spatial-propagation network (SPN) and connected region-aware filter (CRAF). The runtime is calculated on average over all frames and objects.}
		\begin{center}
			\small
			\centering
			\renewcommand{\arraystretch}{1.5}
			\setlength{\tabcolsep}{4pt}
			\begin{tabular}{|l|cccc|}
				\hline
				step       & FS & IR & SPN & CRAF\\           
				\hline
				test time  &0.3s&0.3s&0.08s&0.1s\\
				\hline
			\end{tabular}
		\end{center}
		
		\label{tab:runtime}
		\vspace{-4mm}
	\end{table}
	
	\begin{table*}[t]\scriptsize
		\caption{Overall results on the DAVIS 2017 Challenge}
		\begin{center}
			
			\small
			\centering
			\renewcommand{\arraystretch}{1.3}
			\setlength{\tabcolsep}{2.2pt}
			\begin{tabular}{|l|cccccccccc|}
				\hline
				
				Method      & Ours (cjc) & lixx & apdata  &vantam299& haamooon& voigtlaender & lalalafine123 &YXLKJ & wasidennis & Fromandtozh \\
				\hline
				Global Mean      &0.569 & 0.699& 0.678 &0.638 & 0.615 &0.577&0.569 &0.558&0.548&0.539\\
				J mean           &0.536 & 0.679& 0.651 &0.615 & 0.598 &0.548&0.548 &0.538&0.516&0.507\\
				F Mean           &0.602 & 0.729& 0.706 &0.662 & 0.632 &0.605&0.591 &0.578&0.579&0.571\\
				\hline
			\end{tabular}
		\end{center}
		
		\label{tab:results}
		\vspace{-6mm}
	\end{table*}
	
	\subsection{Runtime Analysis}
	For the model trained offline, the proposed method runs at the speed of 0.78 seconds per object per frame on a Titan X GPU with 12 GB memory. Detailed analysis in each step is shown in Table \ref{tab:runtime}. When taking the fine-tuning time into account, our system runs at about 10 seconds per frame per object on the DAVIS validation set.
	The table shows that the SPN and CRAF steps improve performance significantly without adding much computational costs.

	\subsection{Results in DAVIS 2017 Challenge}
	In the DAVIS 2017 challenge, we test our method without and with CRAF on the test-challenge set. 
	Without CRAF, the J mean and F mean are 51.6\% and 57.9\%, and they are improved by 2\% and 2.3\% with CRAF.
	The final comparisons of top-10 teams are listed in Table \ref{tab:results}.
	As shown in the table, our method is at the 6th place in the competition.

	\section{Conclusion}
	
	In this work, we propose to use the spatial propagation network (SPN) and connected region-aware filter (CRAF) to refine the instance segmentation in both spatial and temporal domains. We show that on the challenging DAVIS 2017 dataset, the proposed methods achieve competitive performance for multiple instance segmentations in videos. 
	
	{\small
		\bibliographystyle{ieee}
		\bibliography{mybib}
	}
	
\end{document}